\documentclass[conference]{IEEEtran}
\IEEEoverridecommandlockouts
\usepackage{cite}
\usepackage{amsmath,amssymb,amsfonts}
\usepackage{graphicx}
\usepackage{textcomp}
\usepackage{xcolor}
\usepackage{amsmath,amsfonts,amssymb}
\usepackage{algorithm}
\usepackage{algpseudocode}
\usepackage{booktabs}
\usepackage{hyperref}
\usepackage{siunitx}
\usepackage{booktabs}
\usepackage{makecell}

\begin{document}

\title{GEM-Style Constraints for PEFT  with Dual Gradient Projection in LoRA}

\author{\IEEEauthorblockN{Brian Tekmen\IEEEauthorrefmark{1}}
\IEEEauthorblockA{\textit{Department of Computer Science} \\
\textit{University of North Carolina Greensboro}\\
Greensboro, USA \\
betekmen@uncg.edu}
\thanks{\IEEEauthorrefmark{1}Work accepted to the NSF REU Symposium at the 2025 IEEE International Conference on Data Mining (ICDM). Correspondence to: betekmen@uncg.edu.}
\and
\IEEEauthorblockN{Jason Yin}
\IEEEauthorblockA{\textit{Department of Computer Science} \\
\textit{North Carolina State University}\\
Raleigh, USA \\
jyin23@ncsu.edu}
\and
\IEEEauthorblockN{Qianqian Tong}
\IEEEauthorblockA{\textit{Department of Computer Science} \\
\textit{University of North Carolina Greensboro}\\
Greensboro, USA \\
q\_tong@uncg.edu}
}

\maketitle

\begin{abstract}
Full fine-tuning of Large Language Models (LLMs) is computationally costly, motivating Continual Learning (CL) approaches that utilize parameter-efficient adapters. We revisit Gradient Episodic Memory (GEM) within the Low-Rank Adapter (LoRA) subspace and introduce I-GEM: a fixed-budget, GPU-resident dual projected-gradient approximation to GEM's quadratic projection. By constraining non-interference solely within the adapter parameters, I-GEM preserves GEM-like stability with orders-of-magnitude lower mean projection overhead. On a 3-task AG News split with induced domain drift, using GPT-2 (355M) and LoRA ($r=8$), I-GEM matches GEM's average accuracy (within $\sim\!0.04$ pts) and outperforms A-GEM by $\sim\!1.4$ pts. Crucially, it reduces projection time vs.\ GEM by a factor of $\sim\!10^3$. These results suggest that applying GEM constraints in the LoRA subspace is a practical pathway for continual learning at the LLM scale.
\end{abstract}

\section{Introduction}

Current AI systems train models under the assumption that the training data is drawn independently and identically distributed (i.i.d.) from a fixed distribution.\cite{shalev2014understanding, goodfellow2016deep} However, in many real‐world applications, such as robotic control \cite{lesort2019continuallearningroboticsdefinition}\cite{Li2024ContinualPD}, personalized assistants \cite{mazumder2024lifelongcontinuallearningdialogue}, or non‐stationary vision tasks \cite{lomonaco2020continualreinforcementlearning3d}, the data distribution evolves over time, and revisiting past data may be infeasible. 

Continual Learning (CL) addresses this challenge by training over a sequence of tasks, while balancing model's ability to retain prior knowledge with its adaptability to new information. This leads to a stability and plasticity trade-off, which is the central study of CL:  high stability reduces the model's learning ability, whereas high plasticity leads to \textit{catastrophic forgetting} \cite{mccloskey1989catastrophic} \cite{french1999catastrophic}, where newly learned information inadvertently overwrites prior knowledge. 

To solve this issue, parameter‐regularization methods such as Elastic Weight Consolidation (EWC) \cite{Kirkpatrick_2017} \cite{de2021continual} and Synaptic Intelligence (SI) \cite{zenke2017continual} penalize drift away from critical pretrained weights. Replay‐based approaches interleave past examples with new training data \cite{rebuffi2017icarlincrementalclassifierrepresentation}, while optimization methods enforce constraints on the optimizer \cite{wang2024comprehensivesurveycontinuallearning}. Among optimization methods, Gradient Episodic Memory (GEM) \cite{lopezpaz2022gradientepisodicmemorycontinual} stands out for providing theoretical non‐interference guarantees by solving a small quadratic program (QP) \cite{frank1956algorithm} at each step to ensure that updates do not harm stored “memories”.  GEM is attractive due to its theoretical guarantees, but is impractical at LLMs scale for three reasons:
\begin{enumerate}
    \item During the training process, gradients from previous tasks are retained in memory for future use. Considering current AI models with billions of parameters, storing massive gradients is infeasible.
    \item The QP that GEM solves to enforce non-interference is performed at every step, and was originally implemented using an external, CPU-based solver. Modern Parameter-Efficient Fine-Tuning (PEFT) \cite{pmlr-v97-houlsby19a} solutions should eliminate CPU bottlenecks by keeping training on the GPU.
    \item Even though GEM solves the dual of the objective, the time complexity is still dominated by the parameter dimension.
\end{enumerate}

PEFT with Low Rank Adaptation (LoRA) \cite{hu2022lora} reduces fine‐tuning overhead by isolating task‐specific updates into a handful of trainable low‐dimensional matrices, shrinking the parameter footprint. This suggests imposing non-interference constraints in the adapter subspace rather than over all model parameters.

We follow this intuition and revisit GEM in the LoRA adapter subspace. Instead of solving a QP at every step, we replace it with a fixed-budget, dual projected-gradient update with warm starts and Lipschitz-informed step size \cite{nesterov2004introductory, beck2009fista, parikh2014proximal, stellato2020osqp, ferreau2014qpoases}. The resulting projection mechanism approximately enforces non-interference constraints within the adapter parameters and can be easily integrated into standard training pipelines, since it only involves basic matrix–vector operations. Specifically, given the current gradient and the stored gradients from previous tasks, we project the update step onto the set of directions that, to first-order, do not increase the losses of previous tasks. This projection is performed in the adapter parameter space and entirely on-device.

\section{ Problem Formulation}

\paragraph{Continual Learning}
We consider a sequence of tasks $\{\mathcal D_t\}_{t=1}^T$, where each $\mathcal D_t = \{x_{t,i}, y_{t,i}\}_{i=1}^{|\mathcal D|_t}$ may come from a different distribution. Let $f_\theta$ denote a pre-trained model with parameters $\theta \in \mathbb R^p$ and loss $\ell(\cdot, \cdot)$. After observing task $t$, the objective is to improve performance on $\mathcal D_t$ while retaining performance on $\mathcal D_{1:{t-1}}$, mitigating catastrophic forgetting. We write the (expected) per-task risks
\[
\mathcal L_k(\theta) = \mathbb E_{(x,y)\sim \mathcal D_k}[\ell(f_\theta(x),y)]
\]

\paragraph{Replay and gradient constraints}
Replay methods maintain small buffers $\{\mathcal M_k\}_{k<t}$ of past samples. Let  $g = \nabla_\theta \mathcal L_t(\theta)$ be the current gradient and \newline $$g_k = \frac 1 {|\mathcal M_k|}\sum_{(x,y) \in \mathcal M_k}\nabla_\theta \ell(f_\theta(x),y)$$ be the average buffer gradient for task $k$. GEM seeks an update direction $\tilde g$ that (to first-order) does not increase any stored past loss by solving the quadratic projection
\begin{equation}
    \label{eq:gem-primal-full}
    \min_{\tilde g \in \mathbb R^p}\|\tilde g - g \|_2^2 \quad \text{s.t.} \quad g_k^\top \tilde g \;\ge\; 0,\;\; \forall k < t.
\end{equation}
The update $\theta \gets \theta - \eta\, \tilde g$ enforces first-order non-interference on the buffered samples.

\paragraph{PEFT with LoRA}
To adapt large models efficiently, we freeze the base weights $\theta_0$ and learn low-rank adapters $\phi$ such that $\theta(\phi) = \theta_0 + \Delta\theta(\phi)$
with $|\phi|\ll |\theta|$. At the layer level, LoRA expresses updates as $\Delta W = BA$ with $\Delta W \in \mathbb R^{d \times k}$, $B \in \mathbb R^{d \times r}$, $ A \in \mathbb R^{r \times k}$, $r \ll \min(d,k)$. Training thus proceeds in the adapter subspace: gradients w.r.t\ $\phi$ are given by $g_\phi=\nabla_\phi \mathcal L_t(\theta(\phi))$.

\paragraph{GEM in the LoRA subspace}
We restrict non-interference to adapter parameters. Define the stacked matrix $G \in \mathbb R^{(t-1) \times d_\phi}$ where $d_\phi$ is the dimension of the LoRA parameter vector $\phi$ (i.e. the total number of trainable adapter parameters), and whose $k$-th row is $g_{\phi, k}^\top$ where $g_{\phi, k}$ is the average adapter-gradient over $\mathcal M_k$. Our goal is to find an adapter-space step $\tilde g_\phi$ close to $g_\phi$ that approximately satisfies the replay constraints:
\begin{equation}
    \label{eq:gem-primal-lora}
    \min_{\tilde g_\phi \in \mathbb R^{d_\phi}}\;\tfrac 1 2 \|\tilde g_\phi - g_\phi \|_2^2 \quad \text{s.t.} \quad G\,\tilde g_\phi \; \ge \; 0,
\end{equation}
followed by the update $\phi \gets \phi - \eta\,\tilde g_\phi$.
Problem~\eqref{eq:gem-primal-lora} enforces non-interference approximately in the adapter subspace on the buffered samples.

\paragraph{Dual form and fixed-budget projection}
The Lagrange dual of \eqref{eq:gem-primal-lora} is a non-negative QP over multipliers $\lambda \in \mathbb R_{\ge 0}^{t-1}$:
\begin{equation}
    \label{eq:gem-dual-lora}
    \min_{\lambda \ge 0} \;\; \tfrac 1 2 \, \lambda^\top(GG^\top)\, \lambda \;+\; (G g_\phi)^\top\lambda,
    \qquad
    \tilde g_\phi \;=\; g_\phi + G^\top \lambda^\star.
\end{equation}
Rather than invoke an external QP solver, we optimize \eqref{eq:gem-dual-lora} with projected gradient descent (PGD) \cite{calamai1987projected}, using warm starts and Lipschitz-informed stepsize $\eta_\lambda \le 1/\sigma_{\max}(GG^\top)$. We run a fixed number of $K$ iterations per training step (typically 2-4).
\paragraph{Practical constraints and scope}
Our formulation operates solely in the LoRA adapter space (the base model is frozen), uses first-order constraints on buffered samples, and targets low overhead by avoiding host-device synchronization and external solvers. These choices trade exactness for practicality at LLM scale; empirical results in Section \ref{sec:exp-setup} show that a small $K$ suffices to approach the behavior of exact GEM while reducing projection time by orders of magnitude.

\section{Methods}

\subsection{Revisit GEM-family Methods}
GEM is a continual learning technique where a subset of samples are stored into an episodic memory from each task as they are learned. Each gradient update is evaluated against the stored samples for previous tasks. If the loss increases for any of them, the gradient is projected to the closest one that satisfies all of the constraints, making catastrophic forgetting unlikely. However, obtaining the new gradient involves solving the dual of a QP, which is computationally expensive.

Average Gradient Episodic Memory (A-GEM)\cite{chaudhry2019efficientlifelonglearningagem} was proposed to reduce the runtime cost of GEM. Instead of solving the QP with multiple constraints, A-GEM samples from the episodic memory and computes a single average gradient. By reducing the number of constraints from $t$ to constant, A-GEM simplifies the optimization problem, making the projection step much more efficient. SOFT-GEM relaxes GEM’s hard constraints via hinge penalties for smoother updates~\cite{hu2020gradientepisodicmemorysoft}.  MEGA combines generative replay with gradient constraints to synthesize buffered examples on‐the‐fly~\cite{NEURIPS2020_0b5e29aa}, trading off memory footprint for additional model capacity. Gradient Projection Memory (GPM) maintains a low-rank subspace of gradients from previous tasks using singular value decomposition and projects the current gradient onto the null space of these subspaces, preventing interference with previously learned knowledge \cite{saha2021gradientprojectionmemorycontinual}.
\subsection{I-GEM: Fixed-budget Dual PGD}
To better adapt GEM for modern LLM settings such as LoRA, we propose Iterative Gradient Episodic Memory (I-GEM), an efficient variant that approximates GEM’s QP using a fixed budget gradient descent. We further enhance I-GEM with optimizations such as adaptive learning rates and warm initialization (see Appendix). On our CL-augmented AG News benchmark, I-GEM outperforms GEM in runtime and surpasses A-GEM in accuracy.

We enforce adapter-space non-interference by solving the projection in Eq.~\eqref{eq:gem-primal-lora} and applying the update $\phi \leftarrow \phi - \eta\, \tilde g_\phi$. We optimize the dual in Eq.~\eqref{eq:gem-dual-lora} with a fixed-budget projected gradient descent (PGD), producing $\tilde g_\phi = g_\phi + G^\top \lambda^\star$.

We optimize \eqref{eq:gem-dual-lora}
 with PGD on $\lambda$ using a fixed budget of $K$ iterations:
 \begin{align*}
     \lambda^{(0)} &\gets \text{warm start from previous step or } \textbf{0}\\
     \lambda^{(k+1)} &\gets \max( \lambda^{(k)} - \eta_\lambda \nabla_\lambda \mathcal D(\lambda^{(k)}), 0)
     \quad\\
 \tilde g_\phi &\gets g_\phi + G^\top \lambda^{(K)}, \end{align*}
 where $\nabla_\lambda \mathcal D(\lambda) = (GG^\top)\lambda + G g_\phi.$
 A Lipschitz-related stepsize is $\eta_\lambda \le 1/\sigma_{\max}(GG^\top)$.

 We describe our I-GEM algorithm in Algorithm \ref{alg:igem}.
 \begin{algorithm}[H]
 \caption{I-GEM: Dual PGD projector in the LoRA adapter subspace}
 \label{alg:igem}
 \begin{algorithmic}[1]
     \Require Current adapter gradient $g_\phi \in \mathbb R^{d_\phi}$, matrix $G \in \mathbb R^{(t-1) \times d_\phi}$ with rows $g_{\phi,k}^\top$, warm start $\lambda^{(0)} \!\ge 0$, stepsize $\eta_\lambda$, iterations $K$
    \For{$k=0, \dots, K\!-\!1$}
        \State $u \gets G^\top \lambda^{(k)}$ \Comment{$d_\phi$-vector}
        \State $r \gets G u + G g_\phi$ \Comment{$(t-1)$-vector, gradient of dual}
        \State  $\lambda^{(k+1)} \gets \max(0, \lambda^{(k)} - \eta_\lambda\, r)$ \Comment{project onto $\mathbb R^{(t-1)}_{\ge 0}$}
    \EndFor
    \State $\tilde g_\phi \gets g_\phi + G^\top \lambda ^{(K)}$
    \State \Return $\tilde g_\phi,\ \lambda^{(K)}$
 \end{algorithmic}
 \end{algorithm}
 
Our I-GEM algorithm is inserted between backpropagation and the optimizer step:
 
 \begin{algorithm}[H]
 \caption{One training step with I-GEM (task $t$)}
 \label{alg:train}
 \begin{algorithmic}[1]
     \State Compute loss on current minibatch $\mathcal B_t$ and backprop to obtain $g_\phi$
     \State Build/update $G$ from replay buffers $\{\mathcal M_k\}_{k < t}$ (rows $g_{\phi,k}^\top$)
     \State $\tilde g_\phi, \lambda  \gets \text{I-GEM}(g_\phi, G, \lambda, \eta_\lambda, K)$
     \State Overwrite adapter gradients with $\tilde g_\phi$ and apply optimizer step (e.g. AdamW)
     \State Update replay buffers for task $t$. Carry over $\lambda$ as warm start for next step
 \end{algorithmic}
 \end{algorithm}
 
After the gradients are computed via backpropagation, I-GEM modifies them before they are applied to update the model parameters.
 \subsection{Feasibility of GEM in the LoRA subspace}
 \label{sec:geom}
 Let $J(\phi) := \frac {\partial \theta}{\partial \phi} \in \mathbb R^{d_\theta \times d_\phi}$ denote the Jacobian from the adapter parameters to model parameters. Its columns span the tangent space of the LoRA submanifold in parameter space at $\phi$. For a small adapter step $\Delta \phi \in \mathbb R^{d_\phi}$, the induced parameter-space change is
 $$\Delta\theta \approx J\Delta \phi$$ 
 With full-space gradients $g= \nabla_\theta \mathcal L(\theta(\phi))$, the chain rule gives $g_\phi = \nabla_\phi \mathcal L(\theta(
\phi)) = J(\phi)^\top g$. Similarly, for each past task $k$ we have $g_{\phi,k}=J(\phi)^\top g_k$. The first-order non-increase constraints in parameter space,
$$
g_k^\top \Delta\theta \;\le\; 0 \quad \forall k,
$$
are therefore equivalent (to first order) to the adapter-space constraints
$$
g_{\phi, k}^\top \Delta \phi \;\le\; 0 \quad \forall k.
$$
(with $\Delta\phi = -\eta \tilde g_\phi$, this is enforced by $G\tilde g_\phi \ge 0$).
Projecting steps in the geometry of $\theta$-space induces the metric $H(\phi) := J(\phi)^\top J(\phi)$ on adapter space. Thus, the "closest" adapter step to the target step to some target step $\Delta \phi_0$ is measured by $\|\Delta \phi - \Delta \phi_0\|_{H}^2$. In practice we project gradients instead of steps and avoid forming $H$ explicitly and work on the adapter gradients directly. Stacking one row per task $G \in \mathbb R^{m \times d_\phi}$ with $G_{k}:=g_{\phi,k}^\top$, the adapter-space GEM projection becomes:
$$
\min_{\tilde g_\phi \in \mathbb R^{d_\phi}}\ \tfrac 1 2 \|\tilde g_\phi - g_\phi \|_2^2 \quad \text{s.t.} \quad G\, \tilde g_\phi \ge \ 0,
$$
and we step with $\Delta \phi =-\eta\,\tilde g_\phi$.
When both LoRA matrices are trainable (bilinear reparameterization), $J(\phi)$ varies with $\phi$. The relations above are local but remain the correct ones for enforcing non-interference in the adapter subspace.  If $J(\phi)$ has full column rank (typically $d_\phi \ll d_\theta$), the adapter-space constraints faithfully mirror $\theta$-space to first order. If $J(\phi)$ is mildly ill-conditioned, stabilizers such as row-normalizing $G$ or thin QR decomposition can improve conditioning without changing the constraints. 

\subsection{Implementation Details}
\label{sec:impl}

We maintain small per-task replay buffers $\mathcal M_k$ of examples. For task $k$ the gradient $g_{\phi,k}$ is obtained with one forward/backward pass on $\mathcal M_k$ while keeping base weights frozen, mirroring the standard GEM procedure but confined to the LoRA subspace. Stacking these row-normalized gradients forms the constraint matrix $G \in \mathbb R^{t \times d_\phi}$, whose modest size in the adapter space keeps both computation and memory stable.

To accelerate convergence of the projected gradient descent, we warm-start within a task by carrying the final iterate $\lambda^{(K)}$ across mini-batches and resetting it to zero only at task boundaries. The stepsize is chosen from a Lipschitz estimate of the dual gradient: we set
\[
\eta_\lambda = \frac{c}{\widehat{\sigma}_{\max}(GG^\top)}, \quad c \in [0.5,0.9],
\]
where $\widehat{\sigma}_{\max}(GG^\top)$ is tracked via one to three steps of power iteration, updated intermittently. Because short power iterations tend to underestimate the spectral norm, we favor $c<1$ for stability.

Each projection runs for a small, fixed budget $K \in \{2,\dots,6\}$. To prevent a few “hard” constraints from dominating the dynamics, we normalize the rows of $G$. Computationally, a single dual step amounts to two matrix–vector products, $u = G^\top \lambda$ and $G u$, so the cost is $O(t d_\phi)$ per step and $O(K t d_\phi)$ per projection. The memory footprint is dominated by storing $G$ of size $t \times d_\phi$, which remains small in the LoRA setting.

For context, the exact GEM projection in the full parameter space invokes a QP whose cost scales as $O(t^2 d_\theta)$ in time and $O(t d_\theta)$ in memory per projection. Operating in the adapter subspace replaces $d_\theta$ with $d_\phi \ll d_\theta$ and dispenses with QP solves altogether, yielding substantial reductions in both compute and memory while preserving the spirit of GEM’s non-interference constraints.

\section{Experiments}
\label{sec:exp-setup}
We assess GEM-style projections in the LoRA subspace on a continual version of AG News. All hyperparameters are detailed in Appendix~\ref{app:hyper}. Reproducibility artifacts (configs, seeds, logs, and an environment file) are included in the supplementary material.

\subsection{Datasets and protocol}
We evaluate on the AG News corpus\cite{zhang2015character}, a 4–way news classification benchmark with each item containing a label, title and description. Each label corresponds to the classes \{1:\textsc{World}, 2:\textsc{Sports}, 3:\textsc{Business}, 4:\textsc{Sci/Tech}\}.  The objective follows previous LM-classifier setups \cite{hendrycks2021measuringmassivemultitasklanguage} and tasks the LM to output the corresponding label to the input. To obtain a domain-incremental continual learning setting \cite{van2019three, delange2021continual, lomonaco2017core50}, we construct three experiences by changing the class priors across time. Experiences are built with disjoint rows and then split $80/20$ into train/test with a fixed seed; the order of experiences is shuffled per seed. We report standard CL metrics (AvgAcc, BWT, FWT, Forgetting).

\subsection{Architectures}
The backbone is GPT\,-2 (355M). We freeze all base weights and train LoRA adapters with rank $r{=}8$, $\alpha{=}32$, dropout $0.05$, targeting attention \texttt{c\_attn} and \texttt{c\_proj}, and MLP \texttt{c\_proj}. A 4-way linear head maps the final hidden state to the option index. Max sequence length is 512 tokens.

\subsection{Baselines}
We compare \textbf{GEM}\cite{lopezpaz2022gradientepisodicmemorycontinual}, \textbf{A-GEM}\cite{chaudhry2019efficientlifelonglearningagem}, and our best performing \textbf{I-GEM}. (Appendix
~\ref{app:hyper} for details)
\subsection{Metrics and reporting}
We report final average accuracy (AvgAcc), backward transfer (BWT), forward transfer vs baseline (FWT), average forgetting, and mean projection overhead (measured with synchronized device timers around the projection routine). Formal definitions appear in Appendix~\ref{app:metrics}. Results are averaged over seeds $\{0,2,5,7,11\}$ and presented as mean\,$\pm$\,std.

\section{Results}
\label{sec:results}
\begin{figure}[t]
    \centering
    \includegraphics[width=1\linewidth]{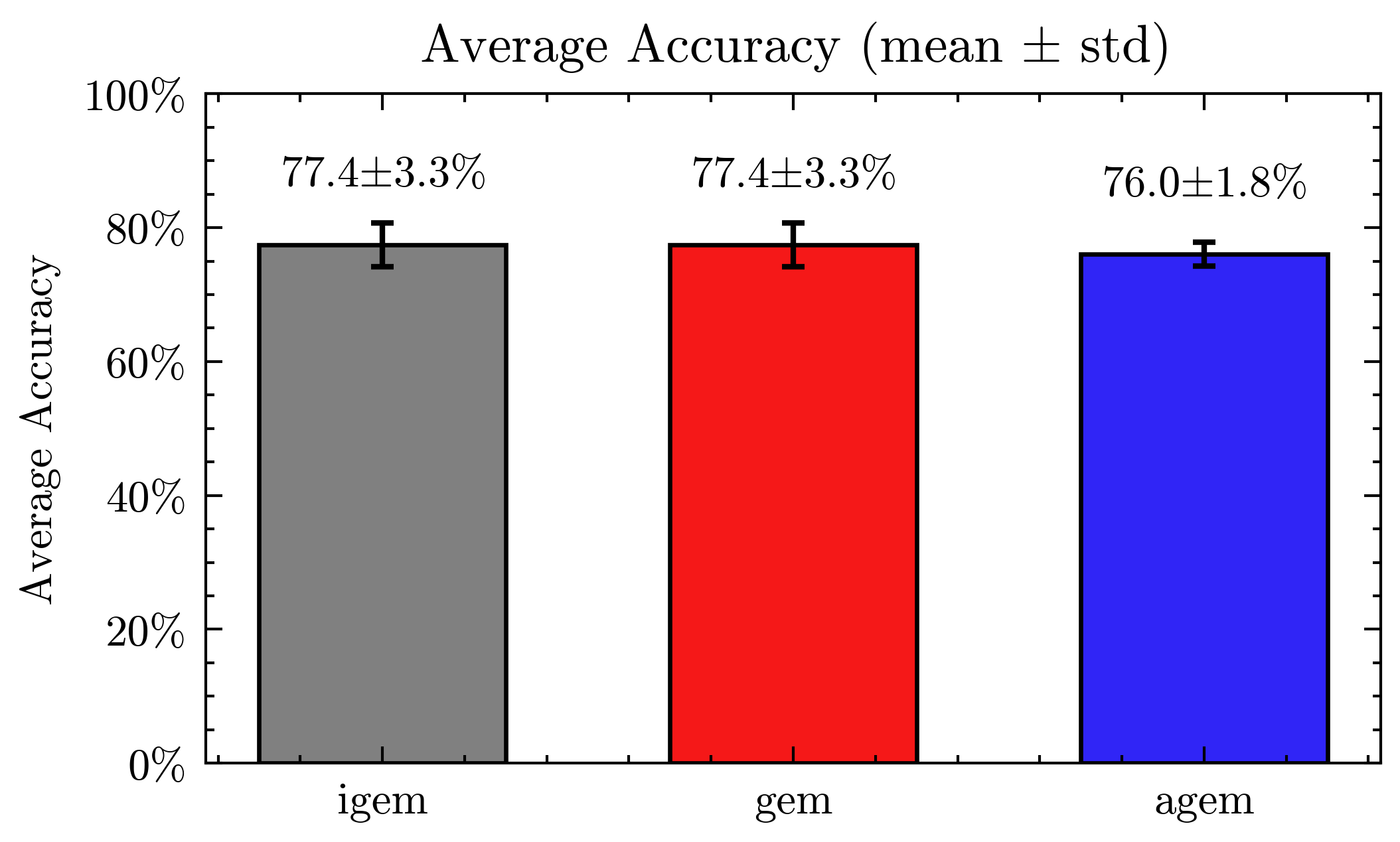}
    \caption{Final Average Accuracy at the end of training.}
    \label{fig:final_avgacc}
\end{figure}

\begin{figure}[t]
    \centering
    \includegraphics[width=1\linewidth]{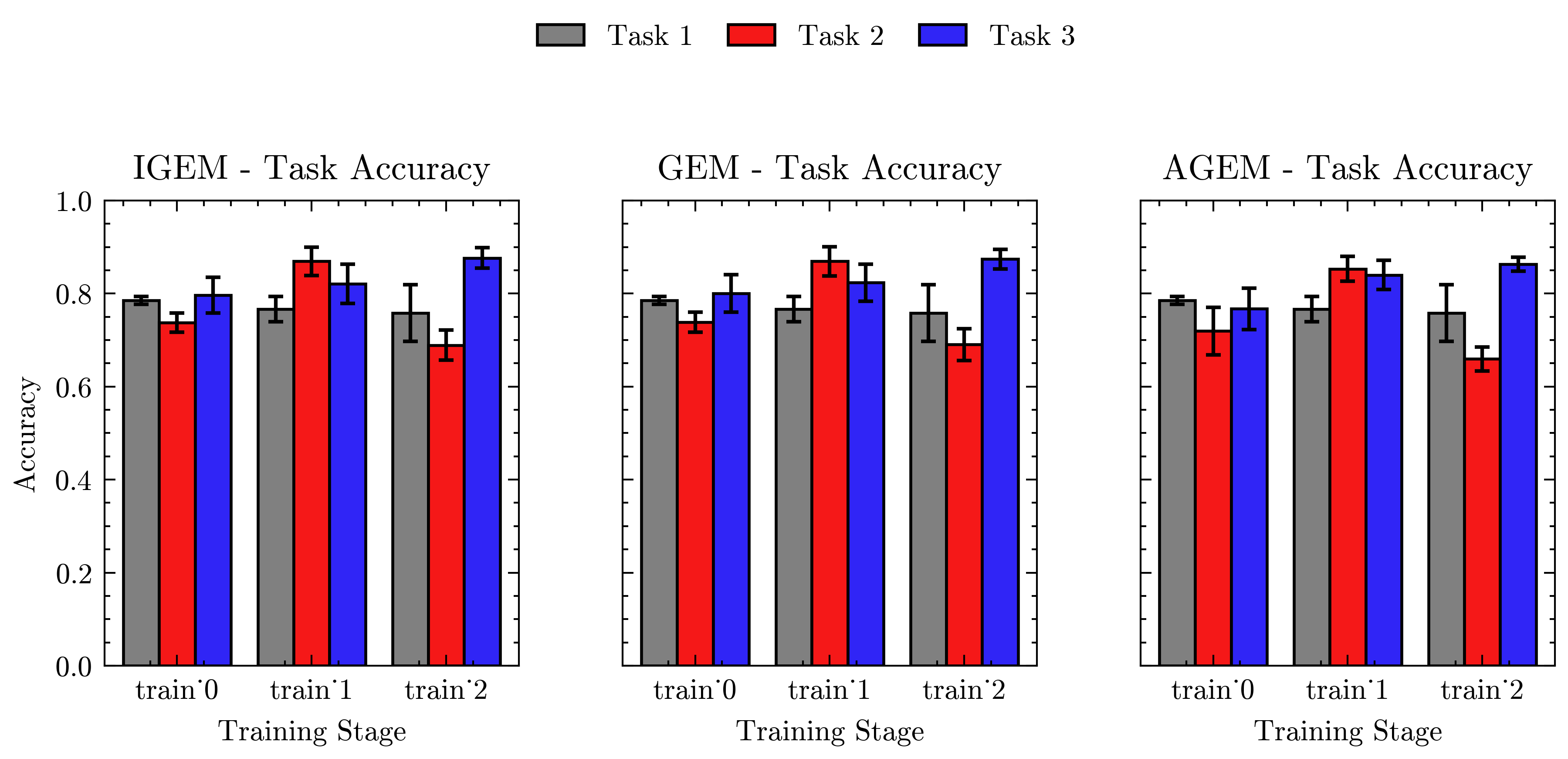}
    \caption{Per\mbox{-}task accuracy vs.\ training step on AG News ($T{=}3$).}
    \label{fig:acc_over_time}
\end{figure}

\begin{figure}[t]
    \centering
    \includegraphics[width=1\linewidth]{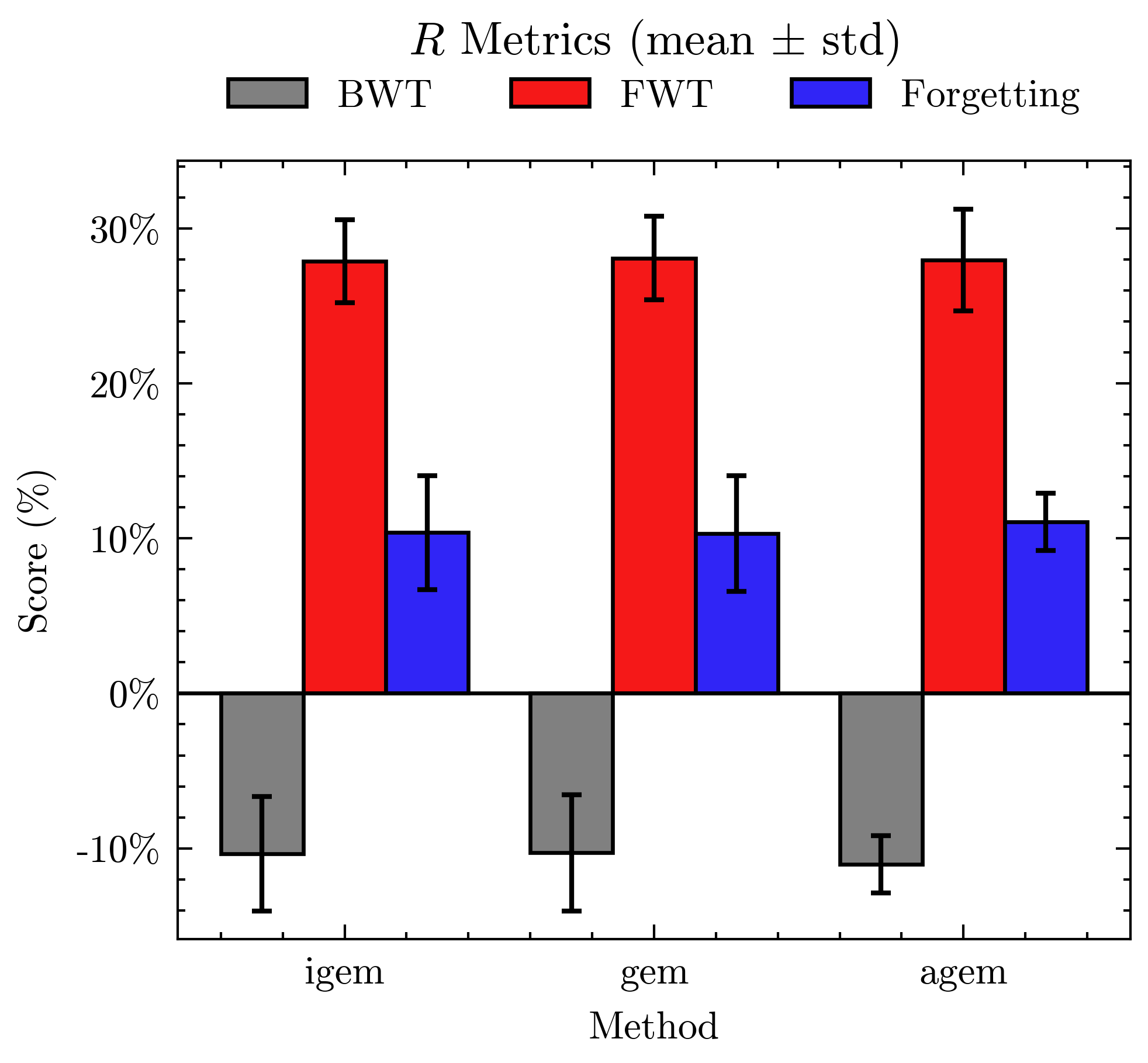}
    \caption{CL metrics (AvgAcc, BWT, FWT) at the end of training.}
    \label{fig:cl_metrics}
\end{figure}

\begin{figure}[t]
    \centering
    \includegraphics[width=1\linewidth]{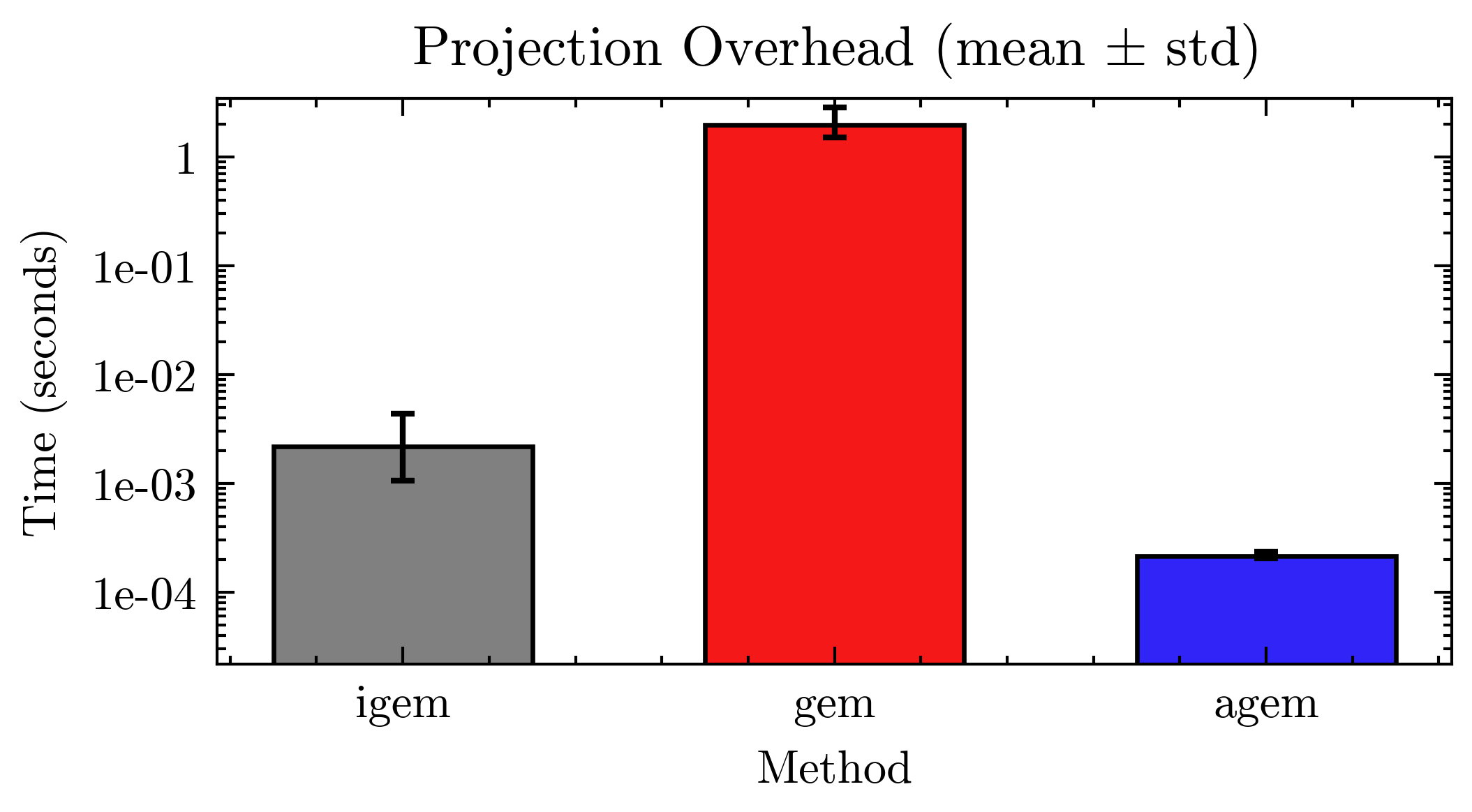}
    \caption{Mean projection overhead (log scale; seconds) measured with synchronized CUDA events.}
    \label{fig:proj_times}
\end{figure}

Table~\ref{tab:avgacc_stats} and Figures~\ref{fig:final_avgacc} and~\ref{fig:proj_times} show that AvgAcc is effectively comparable across methods, with GEM and I\mbox{-}GEM virtually tied and A\mbox{-}GEM within $\sim$1.4 points. I\mbox{-}GEM reduces projection cost by roughly three orders of magnitude vs.\ GEM (mean $0.0022$\,s vs.\ $1.96$\,s), while remaining about $10\times$ slower than A\mbox{-}GEM on projection (Table~\ref{tab:avgacc_stats}). Figure~\ref{fig:acc_over_time} plots per\mbox{-}task accuracies over training. Figure~\ref{fig:cl_metrics} and Table~\ref{tab:cl_metrics} report BWT/FWT/Forgetting; see Sec.~\ref{sec:discussion} for interpretation (e.g., why FWT can appear inflated on AG News).

\begin{table}[t]
\centering
\footnotesize
\setlength{\tabcolsep}{6pt}
\caption{AvgAcc (5 seeds) and mean projection overhead (seconds). AvgAcc reported as mean $\pm$ std.}
\label{tab:avgacc_stats}
\begin{tabular}{lcc}
\toprule
Method & AvgAcc (\%) & Mean Projection Overhead (s) \\
\midrule
GEM    & $77.45 \pm 3.26$ & $1.959 \pm 0.452$ \\
A\mbox{-}GEM  & $76.01 \pm 1.79$ & $2.16{\times}10^{-4} \pm 9.93{\times}10^{-6}$ \\
I\mbox{-}GEM  & $77.41 \pm 3.25$ & $2.16{\times}10^{-3} \pm 1.11{\times}10^{-3}$ \\
\bottomrule
\end{tabular}
\vspace{-4pt}
\end{table}

\begin{table}[t]
\centering
\footnotesize
\setlength{\tabcolsep}{6pt}
\caption{Continual learning metrics (mean $\pm$ std over 5 seeds).}
\label{tab:cl_metrics}
\begin{tabular}{lccc}
\toprule
Method & BWT & FWT & Forgetting \\
\midrule
GEM   & $-0.104 \pm 0.037$ & $0.279 \pm 0.027$ & $0.104 \pm 0.037$ \\
A\mbox{-}GEM & $-0.103 \pm 0.037$ & $0.281 \pm 0.027$ & $0.103 \pm 0.037$ \\
I\mbox{-}GEM & $-0.110 \pm 0.018$ & $0.280 \pm 0.033$ & $0.111 \pm 0.019$ \\
\bottomrule
\end{tabular}
\vspace{-4pt}
\end{table}

I\mbox{-}GEM’s projector uses adapter\mbox{-}space matrix–vector operations and scales per step as $O(K\,t\,d_\phi)$ for $K$ PGD iterations, past tasks $t$, and adapter dimension $d_\phi$, with memory $O(t\,d_\phi)$ for stored past gradients; this keeps the method GPU\mbox{-}resident and explains the large MPO gap vs.\ GEM’s exact QP (Fig.~\ref{fig:proj_times}). While accuracy separation is muted at small $T$, the cost separation is already decisive, with I\mbox{-}GEM $\sim$${\times}\!\!10^3$ faster than GEM and $\sim$${\times}10$ slower than A\mbox{-}GEM per projection.

\section{Discussion}
\label{sec:discussion}

Our results on the CL-augmented AG News domain–drift benchmark show that I-GEM offers the best accuracy–efficiency compromise among GEM-family methods when paired with LoRA. Exact GEM attains strong stability at the price of substantial projection cost; I-GEM preserves GEM-level accuracy (within $\sim\!0.04$ pts) while cutting that cost by $\sim\!×10^3$ via iterative projections in the low-rank adapter subspace. A-GEM remains the fastest method but trails I-GEM in accuracy by $\sim\!1.4$ pts; I-GEM is therefore slower than A-GEM but still efficient in practice, and decisively more accurate.

A central lesson from our study is that the raw AG News corpus is linguistically homogeneous (wire-style boilerplate, recurring named entities, and shared topical phrases) which inflates forward transfer and can mask forgetting. To obtain interpretable continual learning signals without obfuscating the language, we replace naive splits with a simple, reproducible domain shift: each experience uses different class priors (e.g. exp 0 focuses on Sports, exp 1 on Technology, etc.). This protocol replicates the domain-incremental continual learning setting,  zero-shot performance on a new experience is lower, training on the current experience produces a pronounced jump, and retention thereafter is moderate rather than artificially high. Residual forward transfer remains. AG News categories are still semantically related but it is substantially reduced relative to the vanilla setup.

The behavior we observe also highlights that similarity structure, not merely presentation order, governs forgetting and transfer. Experiences that are lexically close tend to stabilize one another, while an outlier domain tends to be forgotten unless replay addresses the mismatch. In this regime, replay buffers should be balanced per label within each task to prevent the classifier head from becoming miscalibrated when class priors change. Finally, while our evaluation demonstrates that I-GEM with LoRA is competitive with full fine-tuning at a fraction of GEM’s projection cost, several avenues remain open: scaling to larger models and longer streams, studying other PEFT and CL approaches, and addressing the privacy footprint of memory by exploring generative or privacy-preserving replay.

\section{Conclusion}
\label{sec:conclusion}

We introduced I-GEM, an iterative projection variant of Gradient Episodic Memory implemented in the LoRA subspace, and evaluated GEM, A-GEM, and I-GEM on an AG News benchmark that induces domain drift by changing class priors across experiences. I-GEM matches the accuracy of exact GEM while drastically reducing mean projection overhead, and it surpasses A-GEM in accuracy while remaining practically efficient despite being slower than A-GEM. The domain–drift protocol clarifies that standard AG News splits overstate forward transfer; under controlled drift, continual-learning behavior is robust and interpretable without altering the language itself. Looking ahead, we expect multi-seed evaluations and memory-size and replay-balance ablations, and studies on larger models and longer streams to further validate the approach. Hybrid methods that combine projection with privacy-aware or generative replay are a promising direction for reducing memory footprints while maintaining stability, positioning I-GEM-style projection in PEFT layers as a practical foundation for real-world continual learning.

\section*{Acknowledgement} 
This work was conducted at the University of North Carolina at Greensboro and supported by the National Science Foundation under the REU program (Award \#2349369). We acknowledge Brian Tekmen and Jason Yin as undergraduate student authors for their contributions.

\bibliographystyle{ieeetr}

\appendix
\subsection{Theoretical Justification}
\noindent In this section, we state the convexity, smoothness, and convergence properties of the dual PGD formulation, and derive error bounds for the approximate projected gradient.

\begin{itemize}
  \item[\bf P1.]  The dual objective 
  \[
    F(\lambda) = \tfrac12\,\lambda^\top (GG^\top)\,\lambda + (Gg_\phi)^\top \lambda
    \quad\text{s.t. }\lambda\ge0
  \]
  is convex over $\mathbb R^T$ and has $L$‐Lipschitz gradient with 
  \[
    L = \sigma_{\max}(GG^\top).
  \]
  \item[\bf P2.]  A projected gradient step with step‐size 
  \(\eta = 1/L\) guarantees
  \[
    F(\lambda^{(k+1)}) \;\le\; F(\lambda^{(k)}) - \tfrac{1}{2L}\|\nabla F(\lambda^{(k)})\|^2,
    \quad\forall k.
  \]
  \item[\bf P3.] Under the above choice of $\eta$, the PGD iterates satisfy
\[
  F(\lambda^{(K)}) - F(\lambda^*) \;\le\;\frac{L\,\|\lambda^{(0)} - \lambda^*\|^2}{2\,K},
\]
i.e.\ an $O(1/K)$ optimality gap. 

\end{itemize}

\bigskip
\noindent Here we derive the full proofs of the above properties.

\subsection{Convexity and Hessian PSD}
\label{app:convexity}
Recall 
\[
F(\lambda)
= \tfrac12\,\lambda^\top (G\,G^\top)\,\lambda \;+\;(Gg)^\top \lambda,
\quad \lambda\ge0,
\]
with Hessian
\[
H\in\mathbb R^{T\times T} := \nabla^2 F(\lambda) = G\,G^\top.
\]
For any $x\in\mathbb R^T$,
\[
x^\top H x = x^\top (G\,G^\top)\,x
= \|G^\top x\|^2 \ge 0.
\]
Hence $H\succeq0$, so $F$ is convex over $\mathbb R^T$ by the Hessian test~\cite{boyd2004convex}.

\subsection{Lipschitz Smoothness of $\nabla F$}
\label{app:lipschitz}
We have 
\[
\nabla F(\lambda) = H\,\lambda + Gg,\quad
H = G\,G^\top.
\]
Since $H$ is constant and symmetric PSD, its operator norm is 
\[
\|H\|_2 = \sigma_{\max}(H), \quad L := \|H\|_2.
\]
 Then for any $u,v\in\mathbb R^T$,
\[
\|\nabla F(v) - \nabla F(u)\|
= \|H(v-u)\|
\le \|H\|_2 \,\|v-u\|
= L\,\|v-u\|.
\]
Thus $\nabla F$ is $L$-Lipschitz continuous~\cite{nesterov2004introductory}.

\subsection{Descent Lemma and Optimal Step-Size}
\label{app:descent}
For an $L$-smooth function, the standard descent lemma states \cite{nesterov2004introductory, boyd2004convex}:
\[
F(v')
\le F(v)
+ \langle\nabla F(v),\,v'-v\rangle
+ \tfrac{L}{2}\|v'-v\|^2,
\quad\forall v,v'.
\]
Apply this with the projected gradient step
\[
v^{k+1}
= \Pi_{\ge0}\bigl(v^k - \eta\,\nabla F(v^k)\bigr).
\]
Using firm non-expansiveness of $\Pi_{\ge0}$~\cite{Baushke1996} gives
\[
\langle\nabla F(v^k),\,v^{k+1}-v^k\rangle
\le -\tfrac1\eta\|v^{k+1}-v^k\|^2.
\]
Substitute into the descent lemma:
\begin{align*}
F(v^{k+1})
&\le F(v^k)
- \tfrac1\eta\|v^{k+1}-v^k\|^2
+ \tfrac{L}{2}\|v^{k+1}-v^k\|^2 \\
&= F(v^k)
- \Bigl(\tfrac1\eta - \tfrac{L}{2}\Bigr)\,\|v^{k+1}-v^k\|^2.
\end{align*}
Moreover, non-expansiveness implies
\[
\|v^{k+1}-v^k\|
\le \|(v^k - \eta\nabla F(v^k)) - v^k\|
= \eta\,\|\nabla F(v^k)\|.
\]
Hence
\[
F(v^{k+1})
\le F(v^k)
- \Bigl(\eta - \tfrac{L\eta^2}{2}\Bigr)\,\|\nabla F(v^k)\|^2.
\]
To ensure strict decrease, require
\(\eta - \tfrac{L\eta^2}{2}>0\), i.e.\ \(0<\eta<2/L\).  Optimizing 
\(\eta - \tfrac{L\eta^2}{2}\) over $\eta$ gives the maximizer
\(\eta^* = 1/L\).

\subsection{Convergence Rate of PGD}
\label{app:convergence}
With $\eta = 1/L$, from the descent-lemma bound we get for each $k$,
\[
F(v^k) - F(v^{k+1})
\;\ge\;\tfrac{1}{2L}\,\|\nabla F(v^k)\|^2.
\]
Summing $k=0,\dots,K-1$:
\[
F(v^0) - F(v^K)
= \sum_{t=0}^{K-1} \bigl(F(v^t)-F(v^{t+1})\bigr)
\;\ge\;\sum_{t=0}^{K-1}\tfrac1{2L}\|\nabla F(v^t)\|^2.
\]
Since $v^*$ is the minimizer over $v\ge0$, convexity implies
\(\langle\nabla F(v^*),\,v^0-v^*\rangle\ge0\).  Standard arguments then yield
\[
F(v^K)-F(v^*) \;\le\;\frac{L\,\|v^0 - v^*\|^2}{2K},
\]
i.e.\ $O(1/K)$ suboptimality.

\bigskip
\bigskip
This completes the detailed derivations of all claims.
\bigskip
\subsection{Evaluation Metrics}
\label{app:metrics}
Let \(R_{j,i}\) denote the test accuracy on task \(i\) after training through task \(j\) (\(i,j=1,\dots,T\)), and let \(R_{0,i}\) be the accuracy on task \(i\) before any continual training (using the pretrained base with randomly initialized LoRA adapters).Let \(N_{\mathrm{proj}}\) be the total number of projection calls and \(\tau_k\) the wall-clock time of the \(k\)-th projection (measured with synchronized device timers on NVIDIA H100).  We then compute:

\begin{itemize}
  \item \textbf{Average Accuracy}~\cite{lopezpaz2022gradientepisodicmemorycontinual}:
  \[
    \mathrm{AvgAcc}
    = \frac{1}{T}\sum_{i=1}^{T} R_{T,i}\,,
  \]
  the mean final accuracy across all \(T\) tasks.
\\
  \item \textbf{Backward Transfer (BWT)}~\cite{lopezpaz2022gradientepisodicmemorycontinual}:
  \[
    \mathrm{BWT}
    = \frac{1}{T-1}\sum_{i=1}^{T-1}\bigl(R_{T,i} \;-\; R_{i,i}\bigr)\,,
  \]
  which measures the average change in performance on past tasks after learning the remaining tasks (negative values indicate forgetting).
\\
  \item \textbf{Forward Transfer (FWT)}~\cite{lopezpaz2022gradientepisodicmemorycontinual}:
  \[
    \mathrm{FWT}
    = \frac{1}{T-1}\sum_{i=2}^{T}\bigl(R_{i-1,i} \;-\; \overline{b}_i)\,,
  \]
  ($\overline b_i$ = baseline accuracy at random initialization)
  which quantifies how much learning on previous tasks improves performance on new tasks, relative to baseline.
  \\
  \item \textbf{Forgetting (F)}:
\[
  \mathrm{F}
  = \frac{1}{T-1}\sum_{i=1}^{T-1}
    \Bigl(\max_{1\le t \le T-1} R_{t,i} \;-\; R_{T,i}\Bigr)\,,
\]
the average (over all but the last task) drop from a task’s best pre-final performance to its final performance.
\\
  \item \textbf{Mean Projection Overhead (MPO)}:
  \[
    \mathrm{MPO}
    = \frac{1}{N_{\mathrm{proj}}}\sum_{k=1}^{N_{\mathrm{proj}}} \tau_k\,,
  \]
the mean wall-clock time of a single projection (measured with synchronized device timers).
\end{itemize}
\subsection{Hyperparameters }
\label{app:hyper}
These are the hyperparameters used for the experiments whose results are shown in the paper.
\begin{itemize}
    \item seed: \{0,2,5,7,11\}
    \item proj\_interval: \{1\}
    \item train\_mb\_size: \{32\}
    \item eval\_mb\_size: \{50\}
    \item n\_experiences $(T)$: \{3\}
    \item train\_epochs: \{1\}
    \item lr: \{0.001\}
    \item patterns\_per\_exp: \{100\}
    \item memory\_strength \cite{lopezpaz2022gradientepisodicmemorycontinual}: \{0.3\}
    \item memory\_size: \{150\}
    \item pgd\_iteration: \{3\}
\end{itemize}
\end{document}